\pdfoutput=1

\documentclass[11pt]{article}

\usepackage[]{ACL2023}

\usepackage{times}
\usepackage{latexsym}
\usepackage{booktabs}

\usepackage[T1]{fontenc}

\usepackage[utf8]{inputenc}

\usepackage{microtype}

\usepackage{inconsolata}
\usepackage{graphicx}
\usepackage{amsmath}
\usepackage{amssymb,amsthm}
\usepackage{booktabs}

\usepackage{tikz}
\usepackage{dirtytalk}
\usepackage{comment}
\usepackage{array}
\usepackage{colortbl}
\usepackage{caption}
\usepackage{subcaption}
\usepackage{listings}
\newtheorem{definition}{Definition}[section]

\newcommand{\defeq}{\overset{\mathrm{def}}{=\joinrel=}}
\newcommand{\sartwell}[1]{tb-#1}
\newcommand{\nozick}[1]{j-#1}
\newcommand{\williamson}[1]{g-#1}
\newcommand{\zagzebski}[1]{v-#1}
\newcommand{\austin}[1]{p-#1}

\definecolor{color_immutable1}{RGB}{106,169,79}
\definecolor{color_immutablen}{RGB}{241,194,50}
\definecolor{color_mutable}{RGB}{166,28,0}

\title{Defining Knowledge: Bridging Epistemology and Large Language Models}

\author{Constanza Fierro$^{\dagger}$ \ \ Ruchira Dhar$^{\dagger\ddagger}$ \ \ 
        Filippos Stamatiou$^{\ddagger}$ \\ \textbf{Nicolas Garneau}$^{\dagger}$ \ \ \textbf{Anders S{\o}gaard}$^{\dagger\ddagger}$ \\
        $^{\dagger}$Department of Computer Science, University of Copenhagen \\ 
        $^{\ddagger}$ Center for Philosophy in Artificial Intelligence, University of Copenhagen
        }

\begin{document}
\maketitle
\def\thefootnote{*}\footnotetext{Correspondance: Constanza Fierro <\href{mailto:c.fierro@di.ku.dk}{c.fierro@di.ku.dk}>, Ruchira Dhar <\href{mailto:rudh@di.ku.dk}{rudh@di.ku.dk}>.}\def\thefootnote{\arabic{footnote}}

\begin{abstract}
Knowledge claims are abundant in the literature on large language models (LLMs); but can we say that GPT-4 truly ``knows'' the Earth is round? To address this question, we review standard definitions of knowledge in epistemology and we formalize interpretations applicable to LLMs. In doing so, we identify inconsistencies and gaps in how current NLP research conceptualizes knowledge with respect to epistemological frameworks. Additionally, we conduct a survey of 100 professional philosophers and computer scientists to compare their preferences in knowledge definitions and their views on whether LLMs can really be said to know. Finally, we suggest evaluation protocols for testing knowledge in accordance to the most relevant definitions.
\end{abstract}

\section{Introduction}

NLP researchers have used the term {\em knowledge}~somewhat haphazardly in the context of large language models (LLMs), e.g., discussing ``knowledge contained in language models'' \cite{jiang-etal-2020-know}, their ``knowledge gaps'' \cite{feng2024don}, or how ``LLMs encode knowledge'' \cite{farquhar2023challenges}, and ``model's internal knowledge'' \citep{kassner-etal-2023-language}. \citet{petroni-etal-2019-language} defined an LLM to \textit{know} a fact if it correctly completes a cloze sentence such as ``The capital of Germany is \_\_'', which are typically generated directly from so-called knowledge graphs. Many have evaluated knowledge in this way \citep[][\textit{inter alia}]{jiang-etal-2020-know, paik-etal-2021-world, dai-etal-2022-knowledge, kassner-etal-2020-pretrained, kassner-etal-2021-multilingual, keleg-magdy-2023-dlama}. However, the predictions of semantically equivalent cloze sentences can be inconsistent\footnote{An LLM may predict Berlin in the above, but Hamburg for ``The city which is the capital of Germany is called \_\_''.} \citep{elazar-etal-2021-measuring, kassner-schutze-2020-negated, fierro-sogaard-2022-factual}, leading to question the meaningfulness of knowledge claims. Should we then require an LLM to predict correctly all the paraphrases of a given fact to say it knows it? What about related facts? Can we really say that an LLM knows that `Lionel Messi plays for Inter Miami' if it does not know that `Lionel Messi resides in Miami'? What, then, are sufficient conditions for saying an LLM {\em knows}? Or more generally, can LLMs know {\em anything}? That is: \begin{quote}Can LLMs have {\em bona fide} knowledge? \end{quote}

\begin{figure}[t]
    \centering
    \includegraphics[width=\linewidth]{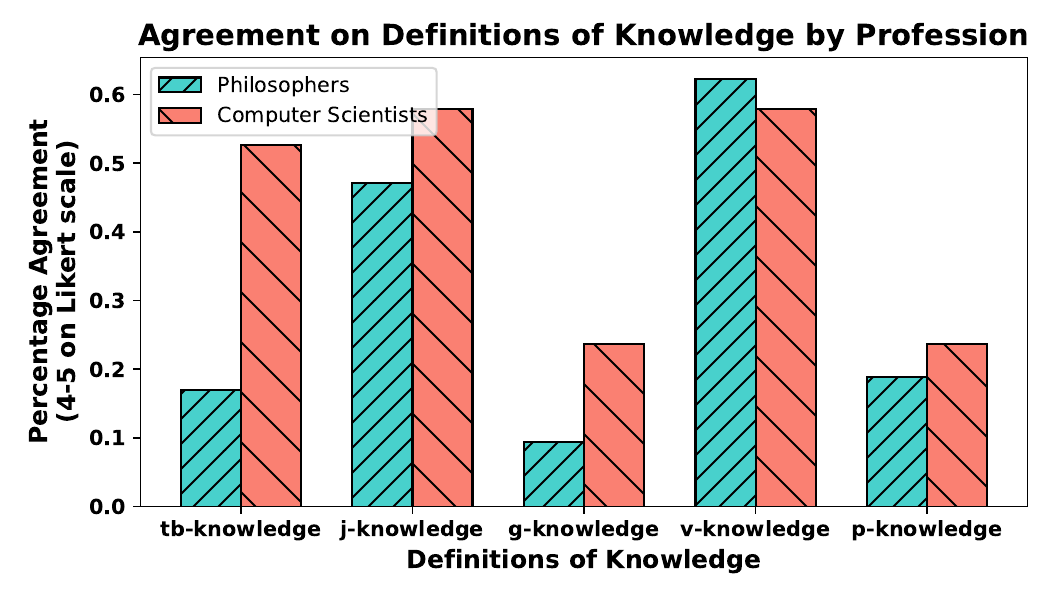}
    \caption{From our survey (\S\ref{sec:survey_results}): Philosophers and computer scientists prefer different definitions of knowledge.
    }
    \label{fig:survey_agreement}
\end{figure}

\begin{table*}[t]
    \centering
    \begin{tabular}{lp{3in}l}
         \toprule &$p$ {\bf is known if and only if}&{\bf Philosopher}  \\
         \midrule 
         {\bf \sartwell{knowledge}}&$p$ is true, and  $p$ is believed$^+$&\citet{Sartwell1992-SARWKI-2} \\   
         {\bf \nozick{knowledge}}&$p$ is true, $p$ is believed, and $p$ is justified&\citet{Nozick2000-NOZKA}\\
        {\bf \williamson{knowledge}}&$p$ is known {\em sui generis} &\citet{Williamson2005-WILKCA}\\
         {\bf \zagzebski{knowledge}}&$p$ is inferred with intellectual virtue&\citet{Zagzebski1999-ZAGWIK}\\
         {\bf \austin{knowledge}}&$p$ is believed and facilitates correct predictions&\citet{Austin2000-AUSOM-3}
         \\ \bottomrule 
    \end{tabular}
    \caption{Five standard definitions of knowledge in philosophy, i.e., knowledge-that $p$ (where $p$ is a proposition). The naming is arbitrary and motivated by keywords. See Appendix \ref{appendix:epistemic_logic} for formalizations in epistemic modal logic. 
    }
    \label{tab:def}
\end{table*}

Whether LLMs know, or in what sense, depends on how \textit{knowing} is defined. Determining what internal knowledge LLMs possess could have important implications on their trustworthiness, as knowledge modulates our trust in agents \cite{Hardwig1991-HARTRO-3,Pederneschi2024-PEDAAO-2}. We tend to lose trust in others when they do not appear to know what we consider basic facts. Furthermore, studying knowledge in LLMs could potentially have implications for epistemology itself~\cite{Cappelen2021-CAPMAI}. 

Recent works have approached the question of how to define knowledge, considering additional requirements for determining what an LLM knows. Some require correct predictions across paraphrases \citep{de-cao-etal-2021-editing, zhong2023mquake}, and others additionally require correct predictions on logically derived facts \citep{kassner-etal-2021-beliefbank, cohen2024evaluating}. However, so far, NLP research has approached knowledge claims in a somewhat arbitrary manner, driven by what seems to make sense intuitively when discussing knowledge. Since philosophy has long tried to define what it means to know, we turn to epistemology to better ground our definitions of knowledge for LLMs.

\paragraph{Contributions} We survey the most commonly used definitions of knowledge in epistemology, and discuss and formalize how to map these definitions to LLMs. We compare current research of knowledge in LLMs to our formal definitions, identifying shortcomings in evaluation practices. We present the results of our survey to philosophers and computer scientists about their views on LLMs and knowledge, finding disagreements about when LLMs can be said to know. These disagreements seem to arise from adherence to slightly different definitions of knowledge (Figure \ref{fig:survey_agreement}). Finally, we provide protocols that follow the epistemological definitions for evaluating and testing knowledge in LLMs. We hope that the connection we provide to epistemology can inform better evaluations and claims regarding knowledge in LLMs.

\section{Definitions of Knowledge}\label{sec:knowledge_defs}

While the NLP research community's use of the word \emph {knowledge} has been somewhat unclear, in philosophy there is a long tradition of trying to pin down exactly what is involved in knowledge claims. Knowledge -- or {\em propositional} knowledge,\footnote{Knowledge is not always propositional; there is also what is referred to as \textit{knowledge-how}, which is related to performance, i.e., knowing how to perform an action \citep{Ryle1949-RYLTCO}.} to be precise -- is what is at stake when we say that `\textit{x} knows that \textit{p}' where \textit{x} is an entity whose knowledge is under question, and \textit{p} is a declarative statement.\footnote{If, for example, $x$= \textit{``John''}  and $p$=\textit{``Berlin is the capital of Germany''}, we can say that $x$ \textit{knows} $p$, if John knows the fact that Berlin is the capital of Germany.} But what are the necessary and sufficient conditions for \emph{knows} here? We review 5 definitions of knowledge (see Table~\ref{tab:def}),\footnote{We have selected five popular epistemological definitions of knowledge, which are among the most common and formal. However, we acknowledge that other perspectives on epistemological knowledge exist. Nonetheless, we believe these five definitions can serve as a solid foundation.} and we interpret and formalize a corresponding definition for LLMs. In \S\ref{sec:knowledge_in_nlp}, we discuss if these definitions are used in the LLM literature, and whether evaluating knowledge claims under them is feasible or not.

\subsection{True beliefs (\sartwell{knowledge})}

\citet{Sartwell1992-SARWKI-2} defines knowledge as a belief that is true, that is `\textit{x} believes that \textit{p}' and `\textit{p} is true'. Mary can on this account believe the capital of Germany is Hamburg, but since Hamburg is {\em not}~the capital of Germany (Berlin is), Mary cannot be said to {\em know} that the capital of Germany is Hamburg. Sartwell argues that there is no need for more requirements for what is knowledge, as long as one has a solid definition of belief. A lucky guess does not qualify as knowledge because, in Sartwell's view, a guess is not a belief. \citet{Sartwell1992-SARWKI-2} requires, in his definition of beliefs, that beliefs are coherent. 
As Sartwell puts it, ``no belief stands in isolation; I cannot have the belief that Goldbach's conjecture is true and fail to have any related beliefs. The belief is constituted as a belief within a system of beliefs.''
Thus we define,

\begin{definition}[belief]\label{def:belief}
    An LLM M believes $p \iff p$ is assigned high confidence.\footnote{This does not simply refer to the output probability assigned to the proposition $p$, as most models could assign fairly high probability to any grammatical sentence, but rather to $M$ assigning high confidence to $p$ relative to other values that $p$ could take.}
\end{definition}

\begin{definition}[belief$^+$]\label{def:belief_plus}
    Let $p, q$ be propositions. A proposition $p$ is believed$^+ \iff $
    \begin{enumerate}
    \item 
$p$ \text{is believed}.
    \item  
$\forall q \;$ st.$\; p\implies q,$ then $q$ is believed.
    \item 
$\nexists q $ st. $q$ is believed $\wedge \, q\implies\neg p$.
    \end{enumerate}
\end{definition}
\noindent That is, $p$ is believed (Def. \ref{def:belief}), any other proposition that follows logically from $p$ is also believed, and $p$ is consistent with any other proposition that is believed (by the same system).\footnote{If I believe in Goldbach's conjecture (any even number greater than two is the sum of two primes), I have to believe the definition of prime numbers, and I can't believe 1+1=3.} 
Thus,

\begin{definition}[\sartwell{knowledge}]\label{def:s_know} An LLM $M$ \sartwell{knows} $p \iff p$ is true $\wedge \; M$ believes$^+$ $p$.\footnote{Our definitions are semi-formal. In epistemic logic, this would be expressed as $\Box_s p\Leftrightarrow p \wedge \Diamond^+ p$. See Appendix \ref{appendix:epistemic_logic}, for epistemic logic formalizations of our knowledge definitions.}
\end{definition}

\subsection{Justification (\nozick{knowledge})}
\citet{Nozick2000-NOZKA} takes another approach and defines knowledge as \textit{justified} true beliefs,\footnote{The idea that knowledge may require some kind of justification goes back at least to Plato \citep[][\textit{187b–201c}]{plato2019theaetetus}. In the Theaetetus, the definition of knowledge as true judgement is ultimately rejected, before  arguing that some sort of account is necessary for knowledge \citep[][\textit{201d-210a}]{plato2019theaetetus}.} with a less strict definition of belief of the sort `$x$ thinks that $p$' and $x$ has some justification for thinking it.\footnote{E.g: Mary thinks there are five oranges on the table, because she counted them up. There really {\em are} five oranges; so Mary knows there are five oranges on the table.}
\citet{Nozick2000-NOZKA} posits that a lucky guess is not knowledge because a guess is not justified. Thus, for LLMs:

\begin{definition}[\nozick{knowledge}] \label{def:n_know} An LLM $M$ \nozick{knows} $p \iff p$ is true $\wedge \; M$ believes $p \; \wedge \; M$ (or $M$'s inference that $p$) is partially interpretable (justified).\footnote{We take this to mean that $M$ can, possibly from ad-hoc methods, provide a rationale for $p$ \cite{joshi-etal-2023-machine}.} 
\end{definition}

\subsection{Sui generis (\williamson{knowledge})}
\citet{Williamson2005-WILKCA} argues for a relativist and primitive view of knowledge, where the truthfulness of $p$ is relative to the agent. Knowledge, on this view, is \textit{sui generis} which is a legal term literally meaning `of its own kind' or `unique'. \citet{Williamson2005-WILKCA} argues that we can't analyze knowledge in terms of other requirements or atomic concepts (belief and justification) because knowledge \textit{is} the atomic concept, which in effect explains what a belief or a justification is and not the other way around.\footnote{In his view, a belief is an attempt at knowing, if I believe the tree in front is a Sequoia then I will act as if I know it. Thus, belief is explain through knowledge and not the reverse.}

\begin{definition}[\williamson{knowledge}] \label{def:w_know} An LLM $M$ \williamson{know} $p \iff M$ includes $p$ in its knowledge bank.
\end{definition} 

We discuss below (\S\ref{sec:knowledge_in_nlp}) what, precisely, it means for propositions to be included in an LLM's knowledge bank. The core intuition is that there is something akin to a knowledge box \cite{Fodor1985-FODFGT} from which known propositions can be extracted. One extreme version would be if the LLM is its own knowledge box, meaning an LLM \williamson{knows} whatever it outputs, but \williamson{knowledge} could also be seen as a modular component in LLM architectures.

\subsection{Virtue (\zagzebski{knowledge})}

The virtue definition of knowledge became popular in the 1980s \citep{Sosa1980-ERNTRA,Greco1993-GREVAV-3}. \citet{Zagzebski1999-ZAGWIK} used it to address the challenge from Gettier cases\footnote{\citet{gettier1963} challenged Nozick's definition of knowledge as (\nozick{knowledge}) by citing a case where justified true belief would not imply knowledge: John sees a sheep in the field and forms the belief that there is a sheep in the field. The sheep that he saw is in fact a dog, but there {\em is} a sheep in the field,  occluded from John's vision. In this case, John had a true belief, as well as a justification (`I saw it with my own eyes') but his justification was false, and John really arrived at the right conclusion out of sheer luck \citep{chisholm1989theory}.} of the justified true belief definition, and states that knowledge is belief arising out of acts of intellectual virtue. As \citet{Zagzebski1999-ZAGWIK} puts it, ``virtues are properties of persons. Intellectual virtues are properties of persons that aim at intellectual goods, most specially the truth.'' An act of virtue is an act in which there is imitation of the behavior of virtuous persons and success in reaching the end for that reason. Therefore if the end is reached by accident and not as a consequence of the virtuous action then it is not considered an act of virtue.\footnote{E.g: A judge determines by an impeccable procedure and motivated by justice that the man is guilty. The judge does everything he ought to do and exhibits all the virtues appropriate in this situation. Nonetheless, for some accidental reason the accused is the wrong man (e.g. the evidence was fabricated). Suppose that the actual killer is secretly switched with the accused man, so the judge ends up sentencing the right man \citep{Zagzebski1999-ZAGWIK}. Here, a feature of luck has cancelled out the bad and the end has been reached, but not because of the virtuous act of the judge.} So we need to define that an LLM is behaving in a virtuous way, that is, it is aiming at the truth and arriving to a prediction as a result of this aim, thus,

\begin{definition}[\zagzebski{knowledge}]\label{def:z_know} An LLM $M$ \zagzebski{knows} $p \iff p$ is true $\wedge \; M$ believes $p \wedge M$'s cause for believing $p$ is motivated only by truthfulness.
\end{definition}

\subsection{Predictive accuracy (\austin{knowledge})}
For \citet{Austin2000-AUSOM-3}, to know means to be able to make correct and relevant assertions about the subject in question. If $M$ \austin{knows} $p$, $M$ believes $p$, and believing $p$ facilitates correct and relevant predictions. Austin's definition is pragmatic. For him ``believing in other persons, in authority and testimony, is an essential part of the act of communicating'', and knowledge is the belief that works out over time. \citet{Austin2000-AUSOM-3} states that knowledge is {\em relevant} true belief under deductive closure; that is, if the subject knows $p$, and believing $p$ implies believing $q$ (with $q$ relevant), then $q$ must be true (and therefore the subject knows $q$ as well). Thus, $p$ facilitates relevant and correct predictions ($q$). This is similar to \sartwell{knowledge}, in which belief$^+$ is epistemically closed, however, in \sartwell{knowledge} the closure scopes over {\em all} propositions $q$, not just the relevant ones. Moreover, since the definition is pragmatic, the deductive closure is only probabilistic. 

\begin{definition}[\austin{knowledge}] \label{def:a_know} Let $p,q$ be relevant propositions st. believing $p \implies$ believing $q$. Then, an LLM $M$ \austin{knows} $p \iff M$ probably \,\sartwell{knows} $p \wedge \, M$ probably \thinspace \sartwell{knows} $q$.
\end{definition}

Relevance is ambiguous and could be defined as $p$ and $q$ being relevant for each other, i.e., $q$ being relevant for knowing $p$; or $p$ and $q$ being relevant for performing a target task (see \S\ref{sec:best_practices}). 

\section{Knowledge in NLP Research}\label{sec:knowledge_in_nlp}

Now, we discuss perspectives from NLP research on what constitutes knowledge, and how these align with the definitions we extracted from the philosophical literature.

\paragraph{\sartwell{knowledge}} Most knowledge probing work seems to rely (loosely) on \sartwell{knowledge} or \austin{knowledge}. Namely, works related to measuring knowledge encoded in LLMs \citep[][\textit{inter alia}]{petroni-etal-2019-language,jiang-etal-2020-know, singh-etal-2020-bertnesia, roberts-etal-2020-much, paik-etal-2021-world, dai-etal-2022-knowledge, kassner-etal-2020-pretrained, kassner-etal-2021-multilingual, dhingra2022time,chalkidis-etal-2023-lexfiles, keleg-magdy-2023-dlama,qi2023crosslingual, fierro2024mulan}, understanding the mechanisms of recalling \citep{dai-etal-2022-knowledge,geva-etal-2023-dissecting,sharma2024locating}, knowledge edits \citep{meng2022locating,hase2023does, meng2023massediting,wang-etal-2024-cross}, and analyses of LLM's knowledge vs contextual factual information \citep{neeman-etal-2023-disentqa, yu-etal-2023-characterizing}. These works follow the LAMA protocol~\citep{petroni-etal-2019-language}, where propositions \(\{p\}\) are derived from knowledge graphs,\footnote{E.g.:\url{https://www.wikidata.org/}} and an LLM is said to know $p$ if it predicts $p$ correctly in a fill-in-the-blank statement. Since $p$ is true (from a knowledge graph) and believed (predicted) by the LLM, the LLM is said to know $p$.\footnote{Note that under this framework we only need to find one surface form of $p$ for which the LLM predicts it correctly to say that it knows $p$.} However, such work fails to address the fact that \sartwell{knowledge} relies on $p$ being believed$^+$, or that \austin{knowledge} requires epistemic closure over relevant propositions.\footnote{Knowledge edits works usually have a mismatch in their definition of knowledge, as they employ true belief (\sartwell{knowledge} without belief$^+$) to determine the set of facts that the model \textit{knows}. But then evaluate the success of an update by measuring correct predictions of paraphrases, and thus accounting to some extent for belief$^+$.} We discuss how best to evaluate whether an LLM believes$^+$ $p$ in \S\ref{sec:best_practices}.

Some works propose to enhance the LLM with an extra component to ensure more consistent beliefs; a so-called {\em belief bank} \citep{kassner-etal-2021-beliefbank} or {\em reflex layer} \citep{kassner-etal-2023-language}. This extra component is optimized for consistency via weighted MaxSAT~\cite{Park2002UsingWM}, and it is used to prompt the model to be consistent to its previous stated beliefs \citep{kassner-etal-2021-beliefbank}, or it is directly used to determine the system's prediction \citep{kassner-etal-2023-language}. Both works aim to rely on \sartwell{knowledge}, where the extra component approximates belief$^+$.\footnote{They track consistency and accuracy to compare systems. Consistency measures the approximation of \sartwell{knowledge}, while accuracy only accounts for belief (Definition \ref{def:belief}).} However, it is only an approximation as the extra component is not necessarily fully consistent and the entailed facts are sampled. This approximation would not be a problem if we consider their approach to be under \austin{knowledge}, although in that case the entailed facts should be selected according to some measure of relevance. 
Furthermore, \citet{kassner-etal-2023-language} are slightly inconsistent in how they use the term knowledge, e.g., using interchangeably ``model beliefs'' and ``models' internal knowledge'', if these were to be the same then they would be talking about \williamson{knowledge}.

\paragraph{\nozick{knowledge}} \citet{hase-etal-2023-methods} adheres to \nozick{knowledge}, but they study LLMs' beliefs and not its knowledge as they argue ``in a traditional view of knowledge as Justified True Belief, it is relatively more difficult to say that an LM knows something rather than believes it''. Nonetheless, they align their experiments with the belief$^+$ definition by measuring beliefs consistency under paraphrasing and entailment.

A justification for \nozick{knowledge} could be provided in different ways, namely, post-hoc attribution to training data using attribution methods \citep{hampel1974influence, koh2017understanding, pruthi2020estimating, akyurek-etal-2022-towards}, logical derivation with a chain-of-thought mechanism \citep{wei2022chain}, generation of factual statements with citations to sources \citep{gao-etal-2023-rarr, menick2022teaching, fierro2024learning}, or potentially as \citet{jiang-etal-2021-know} proposed, the probability of a calibrated language model could be use as justification to differentiate between mere beliefs and knowledge. In any case, the jury is still out on which justification procedures are valid and/or superior, but note that all these methods seem to require partial interpretability.

\paragraph{\williamson{knowledge}} 
One extreme interpretation of the knowledge bank in \williamson{knowledge}'s definition is relativist and deflationary: An LLM knows $p$ if it asserts $p$, simply by generating it. This conflates assertion and true knowledge, and as such, beliefs and knowledge. A more interesting interpretation would be to assume that LLMs have distinct memorization strategies for knowledge and learn to induce modular knowledge components. While some LLM researchers have explored memorization components \citep{dai-etal-2022-knowledge, meng2022locating}, no one has, to the best of our knowledge, identified knowledge components. Some researchers insert devoted knowledge layers \cite{dai-huang-2019-regularization,kassner-etal-2021-beliefbank,kassner-etal-2023-language,feng2024knowledge,liu2024raisf}, which could be interpreted as the knowledge box, but it remains to be seen if such layers permit unambiguous extraction of knowledge claims.

\paragraph{\zagzebski{knowledge}} If knowledge can only be inferred with intellectual virtue, then the difficulty lies identifying intellectual virtues for LLMs. How to test for predictions that are acts of intellectual virtue is an open question. However, we could consider using training data attribution methods as proof of such acts. Another promising avenue is mechanistic interpretability, if we could distinguish factual recall \citep{geva-etal-2023-dissecting} from guessing \citep{stoehr2024localizing} mechanisms. This distinction would relate in interesting ways to the epistemological view of proper functioning \citep{plantinga1993warrant}. \citet{yadkori2024believe} suggest making such a distinction is feasible for some models.

In recent works, \citet{biran2024hopping} address the intellectual virtue condition to some extent by only analyzing the model's virtue knowledge. They do this by filtering out facts $p$ that the model can correctly predict without using critical components in the input, thereby merely guessing the fact (acting unvirtuous). This is a step in the right direction, but a more in depth detection of the inner workings of the model is necessary to filter out all the non-virtuous predictions.

Note that if we interpret the detection of a virtue act can be viewed as a model justification and then it is somewhat unclear what would distinguish \nozick{knowledge} from \zagzebski{knowledge}. This is unsurprising, however, 
since \zagzebski{knowledge} can be seen as an attempt to flesh out what justification turns on \cite{Greco1993-GREVAV-3}. As we insist on concrete methodological interpretations, the two definitions of knowledge may coincide.

\paragraph{\austin{knowledge}} 
In the context of editing factual knowledge in LLMs, \citet{zhong-etal-2023-mquake, cohen2024evaluating} propose to not only evaluate the modified fact itself, but also to evaluate related facts. For example, if we edit an LLM to predict that Lionel Messi now plays in a different football team, then a successful edit should also modify the league in which he plays and the country where he resides. Such evaluation follows the \austin{knowledge} definition, particularly since they focus on evaluating only logically related facts (i.e., only the relevant ones) that are two hops away from the subject or object in question. This type of evaluation could be directly applied to measure the knowledge of the LLM, not just to assess the update accuracy of edits.

The logically related facts to evaluate could also be defined in terms of task relevance. For example, in the context of legal knowledge, \citet{chalkidis-etal-2023-lexfiles} studied the relevance of the knowledge possessed by an LLM for downstream performance in legal classification tasks.

\begin{figure}[t]
    \centering
    \includegraphics[width=\linewidth]{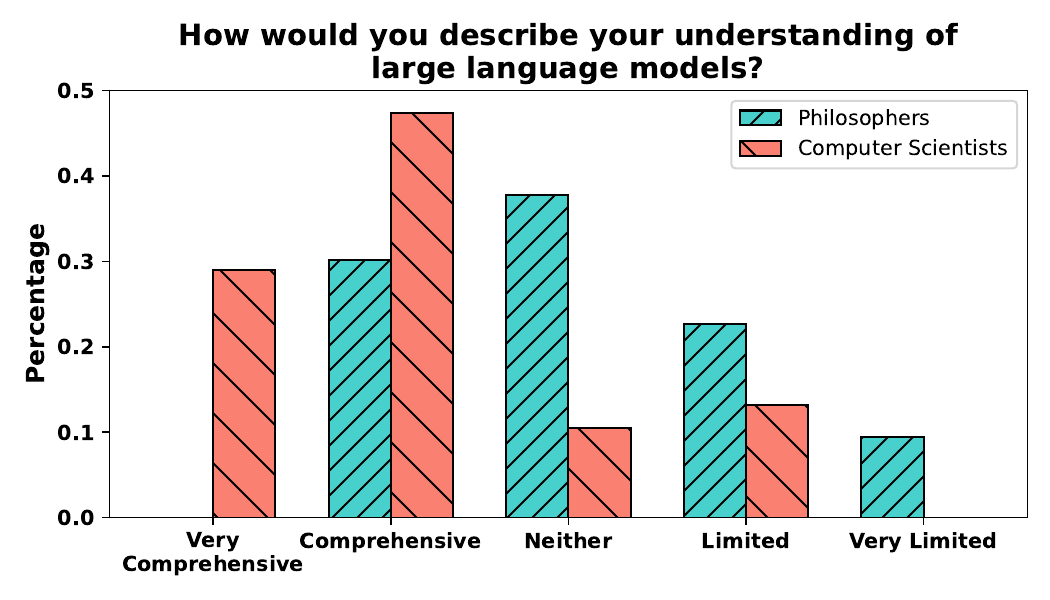}
    \caption{LLMs understanding of respondents.}
    \label{fig:llms_understanding}
\end{figure}

\begin{figure}[t]
    \centering
    \includegraphics[width=\linewidth]{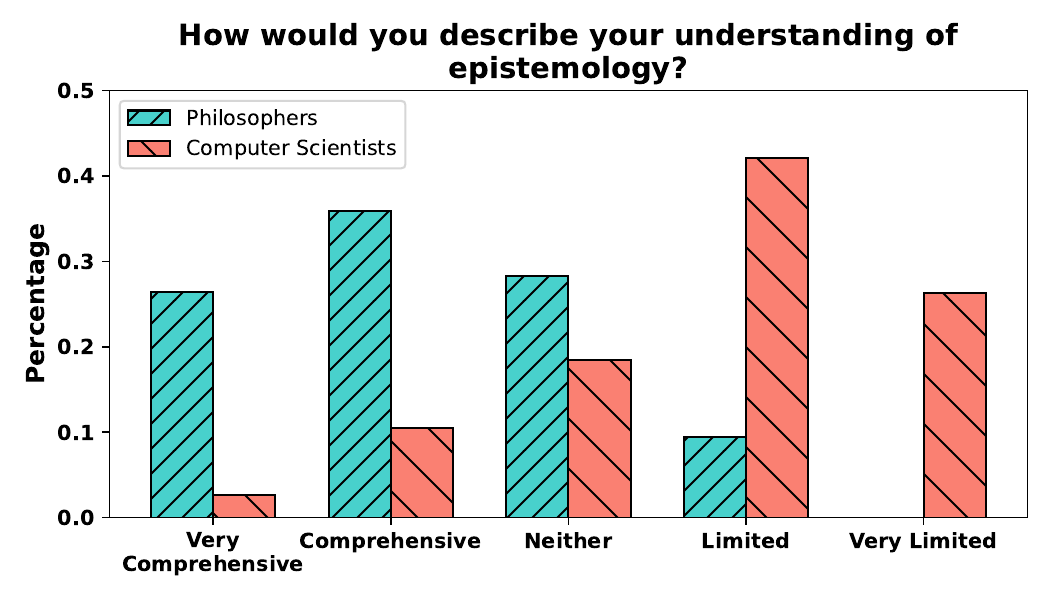}
    \caption{Epistemology understanding of respondents.}
    \label{fig:epist_understanding}
\end{figure}

\begin{figure}[t]
    \centering
    \includegraphics[width=\linewidth]{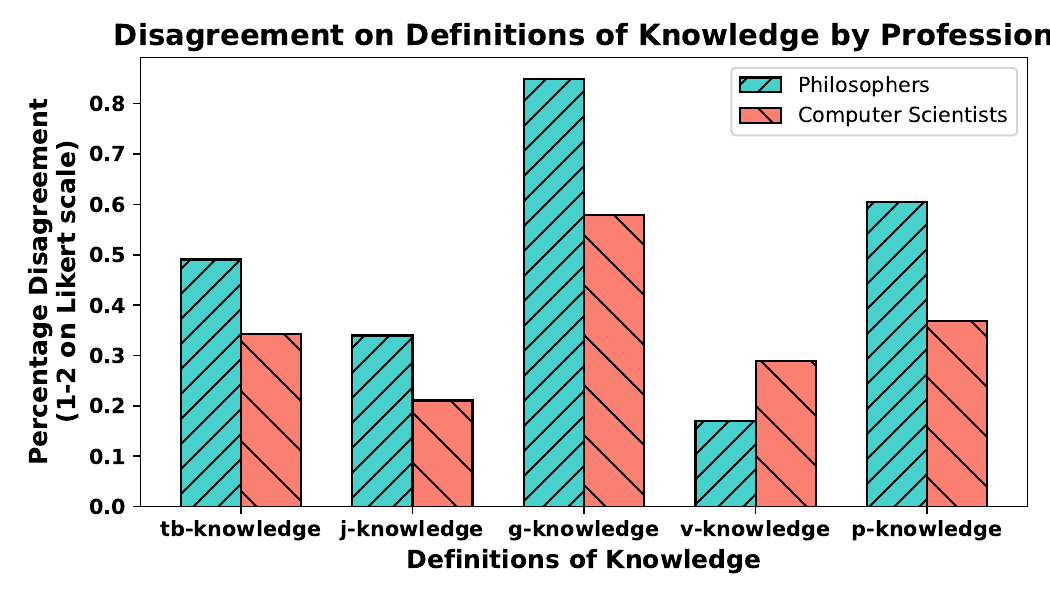}
    \caption{Disagreements on epistemological definitions of knowledge.}
    \label{fig:survey_disagreement}
\end{figure}

\section{Survey Results}\label{sec:survey_results}
To determine how researchers think about knowledge, we turn to our survey of how computer scientists and philosophers. We had 105 respondents, out of which 50.4\% considered themselves philosophers, 36.2\% considered themselves computer scientists, 2.3\% both, and 10.5\% none of the two.\footnote{Some considered themselves mathematicians, cognitive scientists, cultural theorists, etc.} Most respondents from computer science reported a better understanding of LLMs compared to philosophers (see Figure \ref{fig:llms_understanding}) while the majority of philosophers reported better understanding of epistemology compared to 40\% of computer scientists (see Figure \ref{fig:epist_understanding}). See Appendix \ref{appendix:survey_details} for more details. 

\subsection{Questions on Knowledge Definitions}
We asked our respondents to indicate from 1-5 if they disagree completely (1) or agree completely (5) with statements that verbalized our knowledge definitions. See Figure \ref{fig:survey_agreement} and \ref{fig:survey_disagreement} for a summary of the results. In brief, philosophers disagreed with \sartwell{knowledge}, with 49\% selecting 1-2, while the computer scientists agreed more, with 52\% selecting 4-5. Philosophers were divided about \nozick{knowledge},  with a slight tendency to agree (33.9\% chose 1-2 and 47\% chose 4-5). Here, they were in some agreement with computer scientists, 57\% of whom selected 4-5. Philosophers disagreed strongly with the \williamson{knowledge} definition (84\% answers 1-2), whereas computer scientists tended to disagree (57\% answers 1-2). Everyone seemed to like \zagzebski{knowledge} better, with philosophers selecting 4-5 62\% of the time, and computer scientists selecting 4-5 57\% of the time. Philosophers disagreed with \austin{knowledge}, since 60\% selected 1-2; whereas computer scientists seemed more divided, with 36\% choosing 1-2 and 31\% choosing 4-5.

Overall, the survey shows that \nozick{knowledge} and \zagzebski{knowledge} are the most accepted across the two groups.  \sartwell{knowledge} has more mixed results. \footnote{This could either reflect the philosophers' knowledge of the challenges to such definitions of knowledge, or it could reflect the fact that we did not discuss the implications of epistemic closure in the survey (for brevity). In the absence of epistemic closure, maybe some philosophers felt inclined to disagree with this definition.}. The disagreement with \austin{knowledge} is somewhat surprising, since this aligns well with practical evaluation methodologies in the LLM literature.\footnote{One possible explanation was our use of the word  ``useful'' in the survey. This word was intended to convey \austin{knowledge}'s pragmatic flavor, but may have misled some respondents to think that all knowledge has to be directly useful for some user-defined goal.} On the other hand, there is an agreement among philosophers and computer scientists to reject the \williamson{knowledge} definition.

\begin{figure*}
    \centering

\begin{subfigure}[ht]{0.49\linewidth}
    \includegraphics[width=\linewidth]{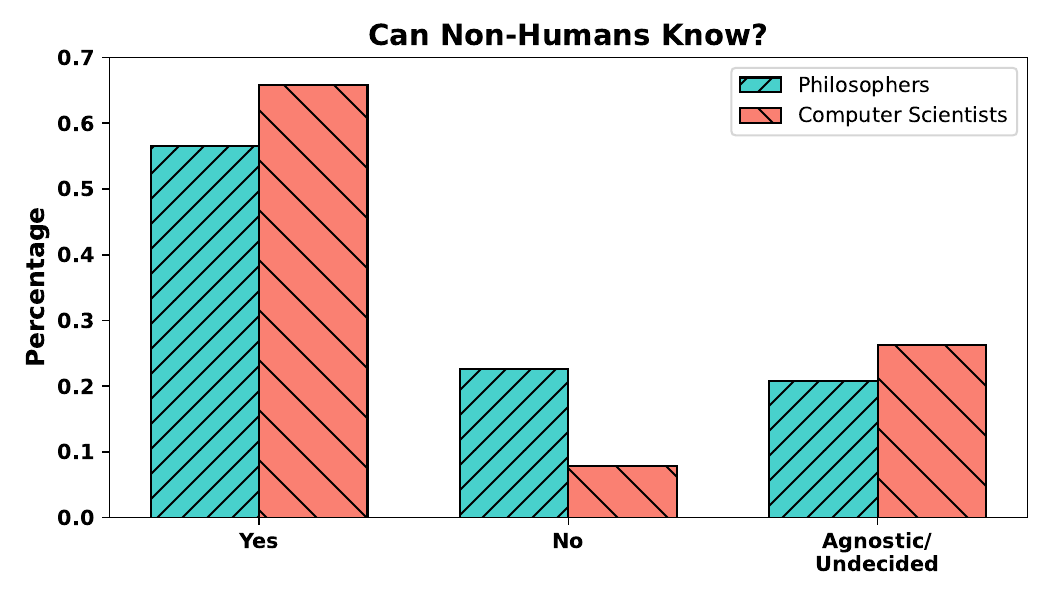}
    \caption{Survey answers to ``Can non-human entities know?''.}
    \label{fig:can_non_humans_know}
\end{subfigure}
\hfill 
\begin{subfigure}[ht]{0.49\linewidth}
    \includegraphics[width=\linewidth]{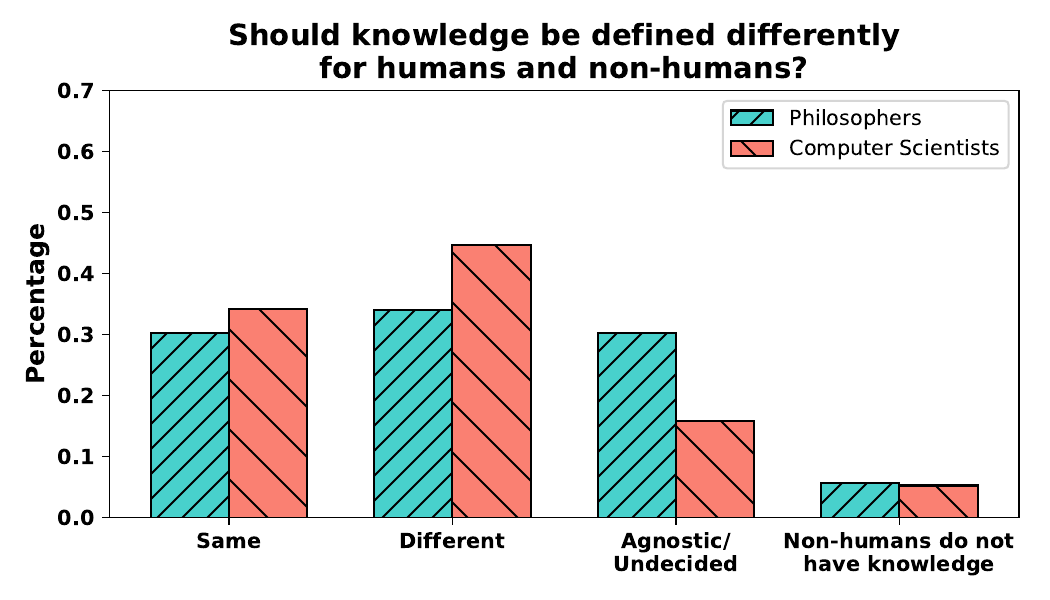}
    \caption{Survey responses on defining global or specific knowledge.}
    \label{fig:same_diff_def}
\end{subfigure}

\begin{subfigure}[t]{0.49\linewidth}
    \centering
    \includegraphics[width=\linewidth]{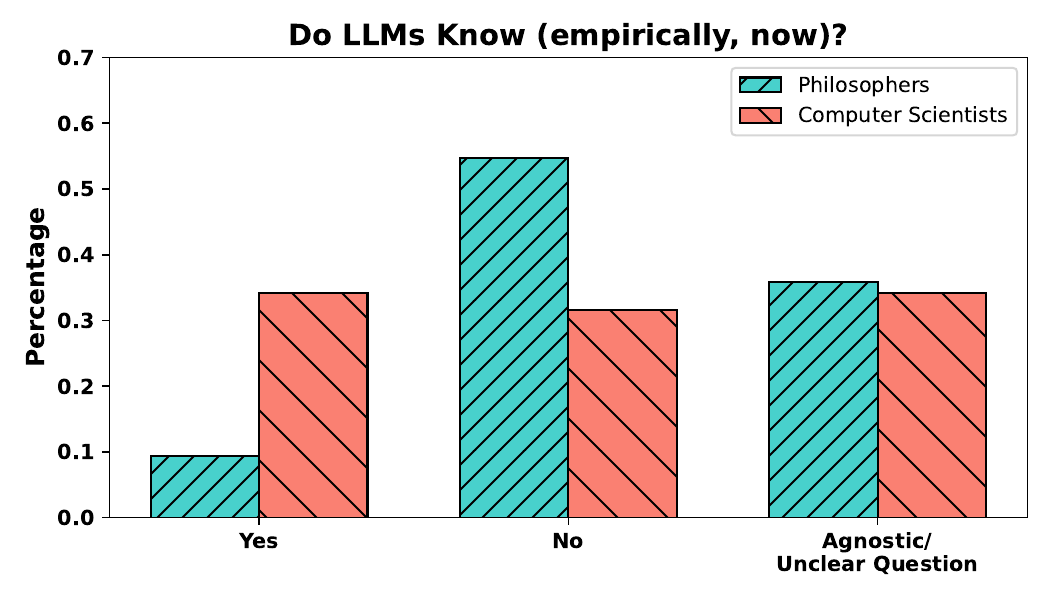}
    \caption{Survey results to the question of LLMs having knowledge.}
    \label{fig:do_llms_know}
\end{subfigure}
\hfill 
\begin{subfigure}[t]{0.49\linewidth}
    \centering
    \includegraphics[width=\linewidth]{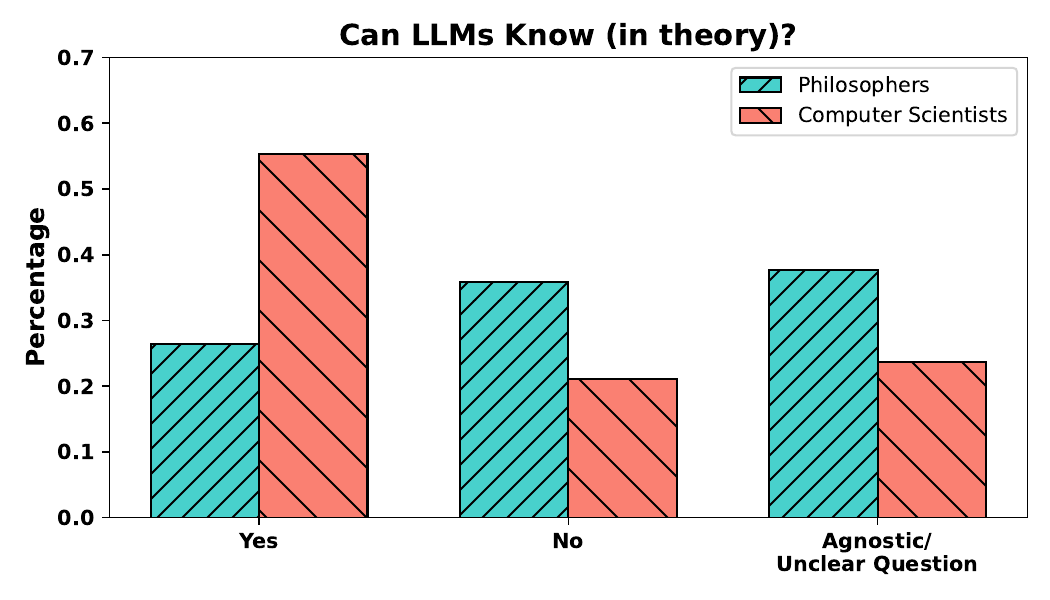}
    \caption{Survey results on LLMs being able to have knowledge.}
    \label{fig:can_llms_know}
\end{subfigure}

\caption{Four of the survey questions and their respective answers.}
\vspace{-3mm}
\end{figure*}

\subsection{General Questions}

\paragraph{Can non-human entities know?} Both computer scientists and philosophers generally agree that non-human entities can possess knowledge (see Figure \ref{fig:can_non_humans_know}). Disagreement within each group is relatively low, with 7\% among computer scientists and 22\% among philosophers.\footnote{This question is intentionally ambiguous, e.g., animals could be consider as non-human entities. We aim to find out whether people think differently about LLMs compared to general non-human entities.}

\paragraph{Should knowledge be defined differently for humans and non-humans?} Computer scientists generally believe that knowledge should be defined differently for humans and non-humans, while philosophers are more divided. Among philosophers, 33\% think it should be different, and 30\% think it should be the same. Among computer scientists, 44\% think it should be different, and 34\% think it should be the same (see Figure \ref{fig:same_diff_def}).

\paragraph{Do LLMs know (empirically, in practice, now)?} There is a significant difference in opinion between philosophers and computer scientists. Philosophers largely disagree, with 54\% saying no and only 11\% saying yes. In contrast, computer scientists are more divided, with 31\% saying no, 34\% saying yes, and the remaining respondents undecided or unclear (see Figure \ref{fig:do_llms_know}). Computer scientists, in other words, evaluate LLM knowledge claims more positively. 

\paragraph{Can LLMs know (in theory)?} When considering the question theoretically (as opposed to in practice), approval increases in both groups (see Figure \ref{fig:can_llms_know}). Among philosophers, 24\% now say yes and 33\% say no, showing a more divided opinion. Among computer scientists, 55\% say yes and 21\% say no, indicating that most believe LLMs can possess knowledge.
\\[12pt]
The survey results thus indicate that scholars from both epistemology and computer science think that the notion of knowledge for LLMs is not a trivial one. Despite differences in opinion, two key points emerge: most scholars believe non-humans can possess knowledge, and LLMs have the potential to "know" in some sense.

\section{Best Practices}\label{sec:best_practices}

Given our discussion of mapping knowledge definitions to LLMs and the results of our survey, we provide possible protocols for evaluating knowledge of LLMs in relation to each discussed definition.\footnote{We provide practical examples on how the definitions could be implemented with the current research. However these protocols may change completely in the future as we better understand the inner workings of LLMs and develop new methodologies and algorithms.} We also provide a really simple example to contrast in a more practical manner some of the definitions. We use Llama-3-8B-Instruct\footnote{\url{https://github.com/meta-llama/llama3}} with greedy decoding for generating completions.\footnote{We use the system prompt: ``You are a helpful chatbot that aims to be truthful.''}

\paragraph{Protocol for \sartwell{knowledge}}
A protocol for evaluating knowledge of $p$ as per Definition \ref{def:s_know} would involve evaluating the three conditions for belief$^+$ (Definition~\ref{def:belief_plus}), which can be done by evaluating model confidence in the true statement itself, as well as in all that follows logically from the true statement. The model should, of course, have low confidence in statements that could imply $\neg p$.

Most current work (\S\ref{sec:knowledge_in_nlp}) evaluates model confidence in $p$, but to assert \sartwell{knowledge} in LLMs, we must also evaluate model confidence in all that is implied by $p$. In our small example (Table \ref{tab:protocol_example}), we evaluate whether Llama-3 knows

\begin{itemize}
    \item[] $p$ = `Platypuses are mammals'
\end{itemize} We first test model confidence in the answer to `Are platypuses mammals?' being {\em yes}. We then evaluate the epistemic closure by evaluating model confidence in facts that follow logically from the platypuses being a mammal, e.g., `Do platypuses have hair or fur?' For this question, the model has more confidence in the answer {\em yes, they have fur}. We now  prompt the model `Do mammals lay eggs?', and the model answers {\em no}. Its answer to `Do platypus lay eggs?' is {\em yes}. Therefore, the model believes 

\begin{itemize}
	\item[]  $q$ = `Platypuses lay eggs and mammals do not'
\end{itemize} 

\noindent which implies $\neg p$, thus violating condition 3 from the belief$^+$ definition; leading us to conclude that Llama-3 does not \sartwell{know} $p$.\footnote{In this example conditions (2) and (3) have been tested with only one proposition that follows logically, but in reality one should obviously sample from a large enough set of propositions. We have also used greedy decoding but different approaches to \textit{high confidence} can be used.}

\paragraph{Protocol for \nozick{knowledge}} If we subscribe to \nozick{knowledge} -- which many computer scientists do (\S\ref{sec:survey_results}) -- then we need to have a two part protocol: (1) Same as in \sartwell{knowledge} the model's confidence in the true statement should be high; and (2) we must also attribute this belief to a training data which unambiguously states $p$, or reasoning that justifies how $p$ can be derived from already established propositions.\footnote{See \S\ref{sec:knowledge_in_nlp} for references to current methodologies of reasoning and training data attribution.}

In our running example, we obtain a justification by prompting Llama-3 with `Are platypus mammals? Please explain step-by-step', for which the model generates the definition of a mammal, platypus characteristics corresponding to mammals' features, and explains that platypus are mammals even though they do not comply with all the mammals' features (exact answer in Appendix \ref{appendix:protocol_example}). By establishing that the intermediate reasoning steps are correct (the characteristics of mammals and platypus) we can conclude that Llama-3 \nozick{knows} $p$.\footnote{We have used chain-of-thought prompting in this example, however it should be noted that the reasoning steps need to be verified for this to be a valid justification \citep{golovneva2023roscoe,jacovi2024chain}.}

\paragraph{Protocol for \williamson{knowledge}}  If by \williamson{knowing} $p$ we simply mean the ability to state $p$, then \williamson{knowledge} will not do much work for us. On such an account, knowledge becomes indistinguishable from beliefs. In line with our discussion in \S\ref{sec:knowledge_in_nlp}, we generally recommend to adopt other definitions.

\paragraph{Protocol for \zagzebski{knowledge}} The \zagzebski{knowledge} definition seems to be quite popular among both philosophers and computer scientists. In \S\ref{sec:knowledge_in_nlp}, we cited possible interpretations of intellectual virtue in LLMs. Training data reliability assessments could involve attributing the inference of $p$ to training data that contains $p$, and showing that the model knows this data is reliable, 
e.g., by using a linear probe to see whether the model successfully distinguishes reliable from unreliable training data. On the other hand, if the model infers $p$ from in-context data that we know is reliable, we need to show that the model is indeed generating the proposition using the provided in-context knowledge, e.g., via mechanistic interpretability \citep{yu-etal-2023-characterizing, wu2024retrieval}.

\paragraph{Protocol for \austin{knowledge}} 
If knowledge is something that facilitates correct predictions, we need to be able to sample from the set of relevant situations. This is of course a familiar challenge to LLM researchers interested in evaluating performance in the wild. We propose to evaluate \austin{knowledge} as we would evaluate \sartwell{knowledge}, albeit in a probabilistic setting, and only over the relevant set of implied propositions.\footnote{This seems to make the \austin{knowledge} definition {\em strictly weaker} than \sartwell{knowledge}, with the implication that any model that \sartwell{knows} $p$ will also \austin{know} $p$. This conclusion depends on whether our notion of model usefulness is limited to knowledge. If we can dissociate knowledge performance from task performance and talk about model usefulness only in terms of knowledge, it holds that \austin{knowledge} is strictly weaker than \sartwell{knowledge}. If not, we must add the additional requirement that models perform well on the domain they are supposed to be knowledgeable about.}  While computer scientists prefer \sartwell{knowledge} over \austin{knowledge} (by some margin; see \S\ref{sec:survey_results}), the definition of \austin{knowledge} seems more in line with current practices in the LLM community.
Following with the example in Table \ref{tab:protocol_example}, here, 
we would conclude that Llama-3 \austin{knows} `Platypuses are mammals', as opposed to \sartwell{knowing}. Since even though believing mammals do not lay eggs, is in contradiction with $p$, $q$ is true \textit{most of the times}.

\section{Conclusion}
In this paper, we reviewed epistemological definitions and formalized interpretations in the context of large language models (LLMs). Then, we examined how existing works in NLP research align with these definitions, highlighting gaps in their interpretations of knowledge. Furthermore, we presented the results of our survey of philosophers and computer scientists, showcasing the different views in terms of definitions of knowledge and whether LLMs can be said to know. Finally, we outlined protocols of evaluations for each knowledge definition using existing algorithms and methodologies. We hope that the connection to epistemological definitions of knowledge can inform the evaluations of knowledge in LLMs and can provide a more solid foundation for the necessary tests to determine when an LLM truly knows a fact.

\section*{Limitations}
We presented five standard definitions of knowledge in philosophy. However, there are more nuances and potentially additional definitions that could apply, nonetheless, we believe these are the most standard and serve as a starting point to ground the evaluations of knowledge in LLMs more formally. Regarding Section \ref{sec:knowledge_in_nlp}, there are certainly more works evaluating knowledge in LLMs that could be included. Nonetheless, we included as many as possible and believe these lay out the current landscape of knowledge evaluation. Finally, as stated in the main body, the protocols are practical methodologies that may become irrelevant as more research on LLMs is conducted. However, we included them here to clarify how the definitions can be implemented in practice.

\section*{Acknowledgements}
We thank our colleagues at the Center for Philosophy in AI and the CoAStaL NLP group for insightful discussions throughout this project. In particular, we would like to thank Daniel Hershcovich, Ilias Chalkidis and Jiaang Li for valuable comments on the final manuscript.  This work has been supported by Carlsberg Semper Ardens Advance Grant CF22-1432.

\bibliography{anthology,custom}
\bibliographystyle{acl_natbib}

\appendix

\section{Epistemic logic}\label{appendix:epistemic_logic}

The syntax of standard epistemic logic is defined by:

$$\phi \defeq p \mid \neg \phi \mid (\phi\wedge\psi)\mid \Box \phi \mid \Diamond \phi $$

The veridicality principle (also known as axiom {\bf T}) that what is known, is also true, is expressed as follows: $\Box\phi\rightarrow \phi$. We will distinguish between different definitions of knowing by subscripting the modal operators. One standard epistemic logic is the so-called {\bf S4} logic, axiomatized as follows: 

\begin{itemize}
    \item[{\bf K}] $\Box(\phi\rightarrow \psi)\rightarrow (\Box\phi\rightarrow \Box\psi)$
    \item[{\bf T}] $\Box \phi \rightarrow \phi$
    \item[{\bf 4}] $\Box\phi\rightarrow \Box\Box\phi$
\end{itemize}

Axiom {\bf 4} is also called the principle of positive introspection. This is not the only epistemic modal logic on the table, but it suffices for our purposes. We extend {\bf S4} in various ways to accommodate for the five definitions. Specifically, \zagzebski{knowledge} introduces the concept of virtue, and \austin{knowledge} relies on some notion of empirical risk. The virtue definition of knowledge introduces a new operator that does not satisfy the veridicality principle {\bf T}. 

\paragraph{\sartwell{knowledge}} A na\"{i}ve implementation of knowledge as true belief falls out of {\bf S4} and the principle called {\bf KB1}, which goes all the way back to Plato: 

$$\Box_s\phi\rightarrow \Diamond\phi$$

\citet{Sartwell1992-SARWKI-2}, however, relies on an extended notion of belief which we will have to formalize, also. Let us introduce a new operator $\Diamond^+$ and call the epistemic closure principle for this operator for {\bf +}: 

\begin{itemize} 
    \item[$\Diamond^+$:] $\Box_s p\rightarrow \Diamond^+ p$
    \item[{\bf +}:] $\Diamond^+ p \rightarrow ((p\rightarrow q)\rightarrow \Diamond^+ q)$
\end{itemize}

One way to express that belief is consistent is by the principle: 

$$\neg\Diamond\bot $$

\paragraph{\nozick{knowledge}} The idea of justified true beliefs calls for so-called {\em justification logic} \cite{Artemov2008-ARTTLO} with justification operators: 

$$\Box_n\phi\rightarrow t:\phi$$

Justification logic can be axiomatized in different ways, but these details go beyond our main concerns here. 

\paragraph{\williamson{knowledge}} If we insist on a {\em sui generis} interpretation of knowledge in {\bf S4}, we would have to introduce a new operator, say $\dagger$. This operator would have very different properties from the standard epistemic modal logic $\Box$-operator. {\bf T} would not apply. {\bf K}~would apply, and knowledge would still be required to be consistent. It is unclear whether {\bf 4} would apply to the $\dagger$-operator. A minimal axiom system could, perhaps, be something like this:

\begin{itemize}
    \item[{\bf T}] $\dagger(\phi\rightarrow\psi)\rightarrow (\dagger\phi\rightarrow \dagger\psi)$ 
    \item[{\bf 0}] $\neg \dagger\bot$
\end{itemize}

We leave further details open. 

\paragraph{\zagzebski{knowledge}} Virtue reliabilist accounts of knowledge and justification are versions of epistemological externalism. Sosa characterizes an intellectual virtue, very generally, as “a quality bound to help maximize one’s surplus of truth over error” (1991: 225). For most virtue reliabilists, intellectual virtue is what leads to justification, and virtue-based knowledge definition are therefore often formalized in justification logics. 

$$\Box_w\phi\rightarrow t:\phi$$

but with slightly different model-theoretic semantics than $\Box_n$. 

\paragraph{\austin{knowledge}} Definition~2.7 translates into the following in an extended probabilistic version of {\bf S4}: 

$$\Box_a p \rightarrow \Diamond p \wedge P(\Diamond q\mid p\rightarrow q )>0.95$$

which is taken to mean that the definition of \austin{know}ing $p$ ($\Box_a p$) is that you believe $p$ ($\Diamond p$), and $q$ follows from $p$, then you probably also believe $q$, i.e., the probability of you believing $q$ ($P(\Diamond q\mid p\rightarrow q )$) is higher than some threshold, say 0.95. This is simply the probabilistic version of \sartwell{knowledge}. The definition of \austin{knowledge} also requires pragmatic usefulness. One way to formalize this is in terms of empirical risk on relevant benchmarks. An alternative is formalization through relevance logics \cite{Urquhart1972-URQSFR}. 

\section{Survey Details}
\label{appendix:survey_details}

We recruited professional LLM researchers and philosophers through Computer Science and Philosophy mailing lists, research groups mailing lists, and point of contacts of the authors at other universities. All respondents participated free of charge on a completely voluntary basis. The respondents were informed about the intended use of the survey. The full instructions was:
\begin{quote}
The <X> is running a survey about the relationship between knowledge and Large Language Models (LLMs). We are interested in getting as many perspectives as possible, especially from philosophers and computer scientists. The survey should not take more than 5 minutes of your time.     
\end{quote}

\begin{figure}[t]
    \centering
    \includegraphics[width=\linewidth]{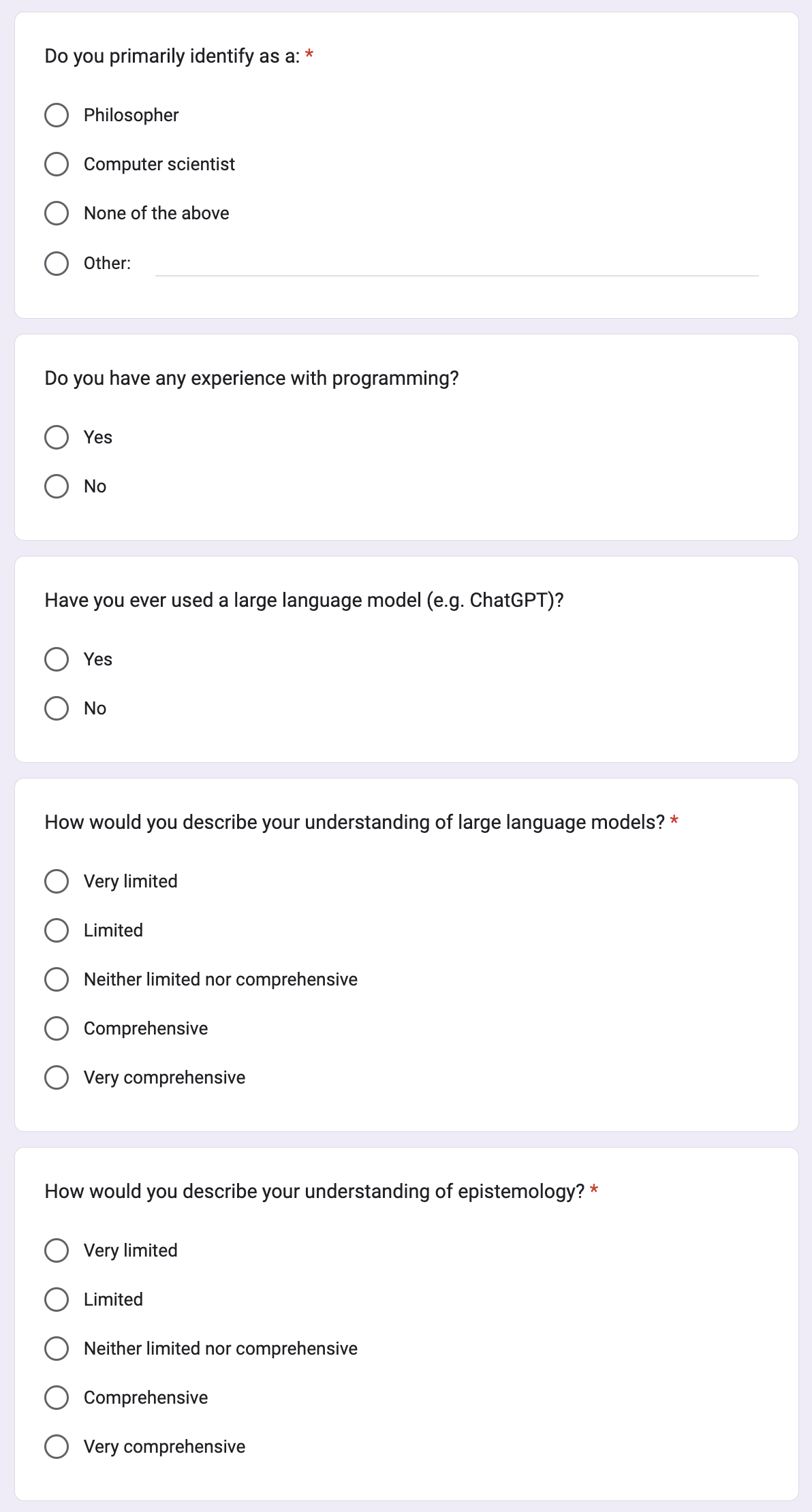}
    \caption{First part of questions from our survey.}
    \label{fig:survey_intro_qs}
\end{figure}

\begin{figure}[t]
    \centering
    \includegraphics[width=\linewidth]{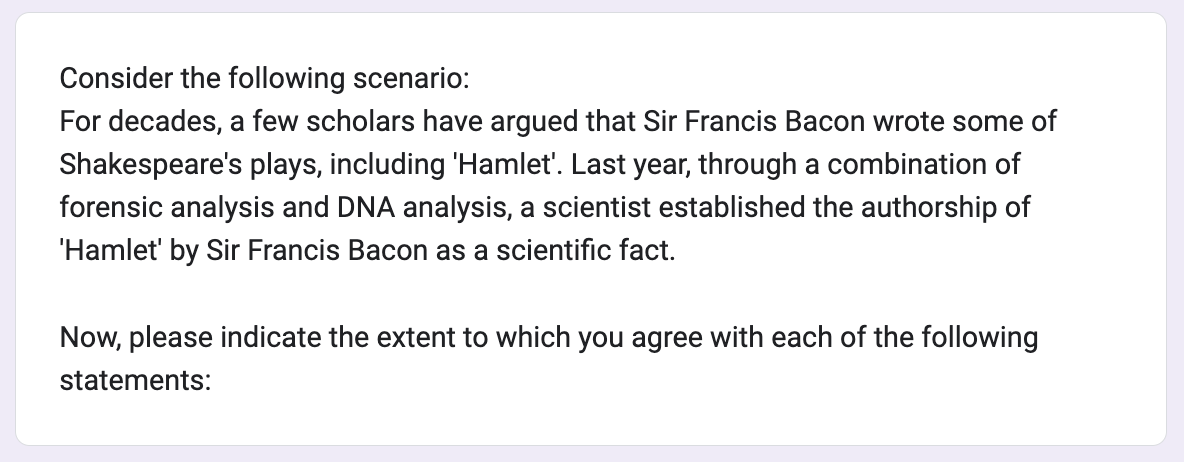}
    \caption{Example to be used for the informal knowledge definitions.}
    \label{fig:survey_scenario}
\end{figure}

\begin{figure}[t]
    \centering
    \includegraphics[width=\linewidth]{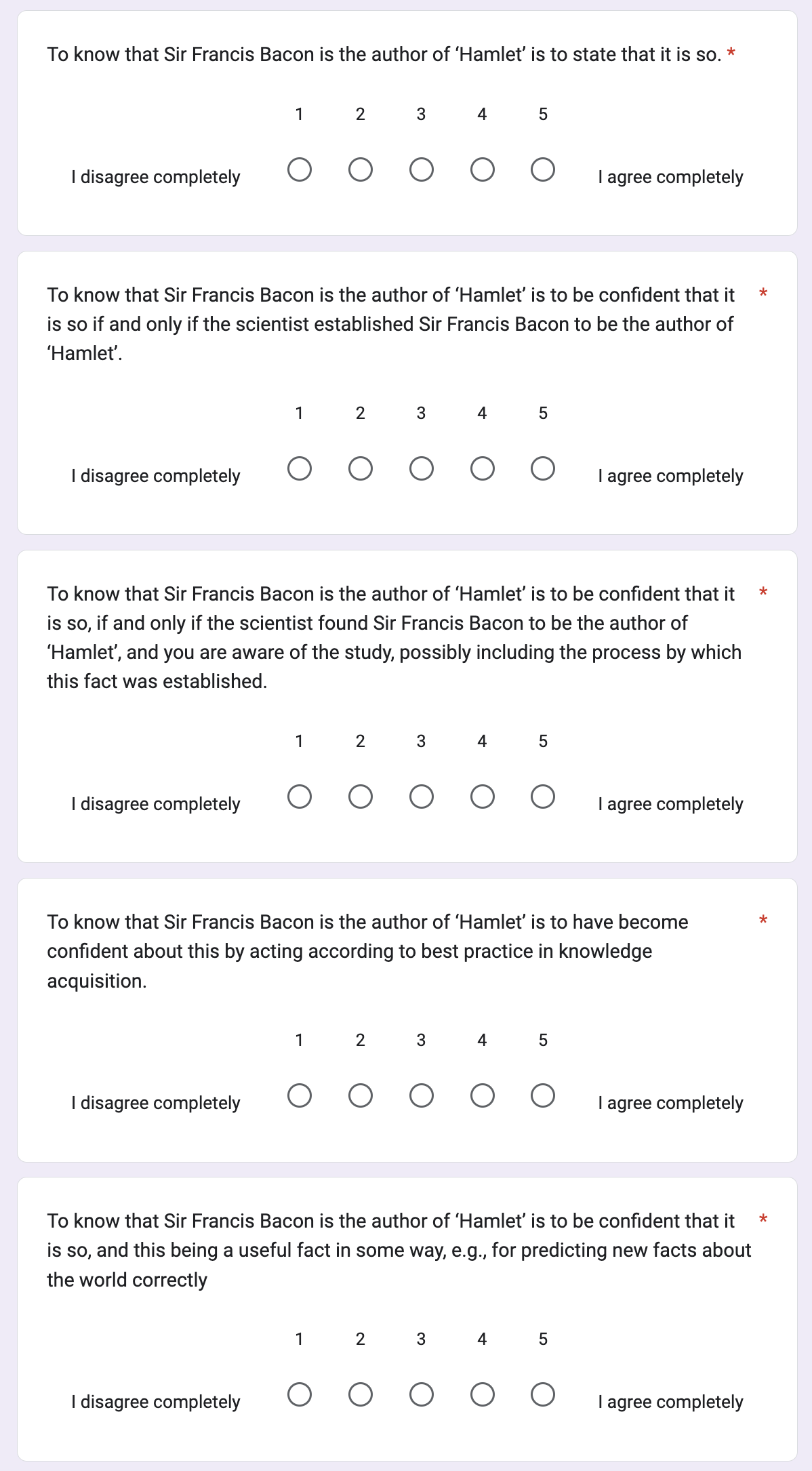}
    \caption{Second part of questions from the survey, related to the knowledge definitions.}
    \label{fig:survey_definitions}
\end{figure}

\begin{figure}[t]
    \centering
    \includegraphics[width=\linewidth]{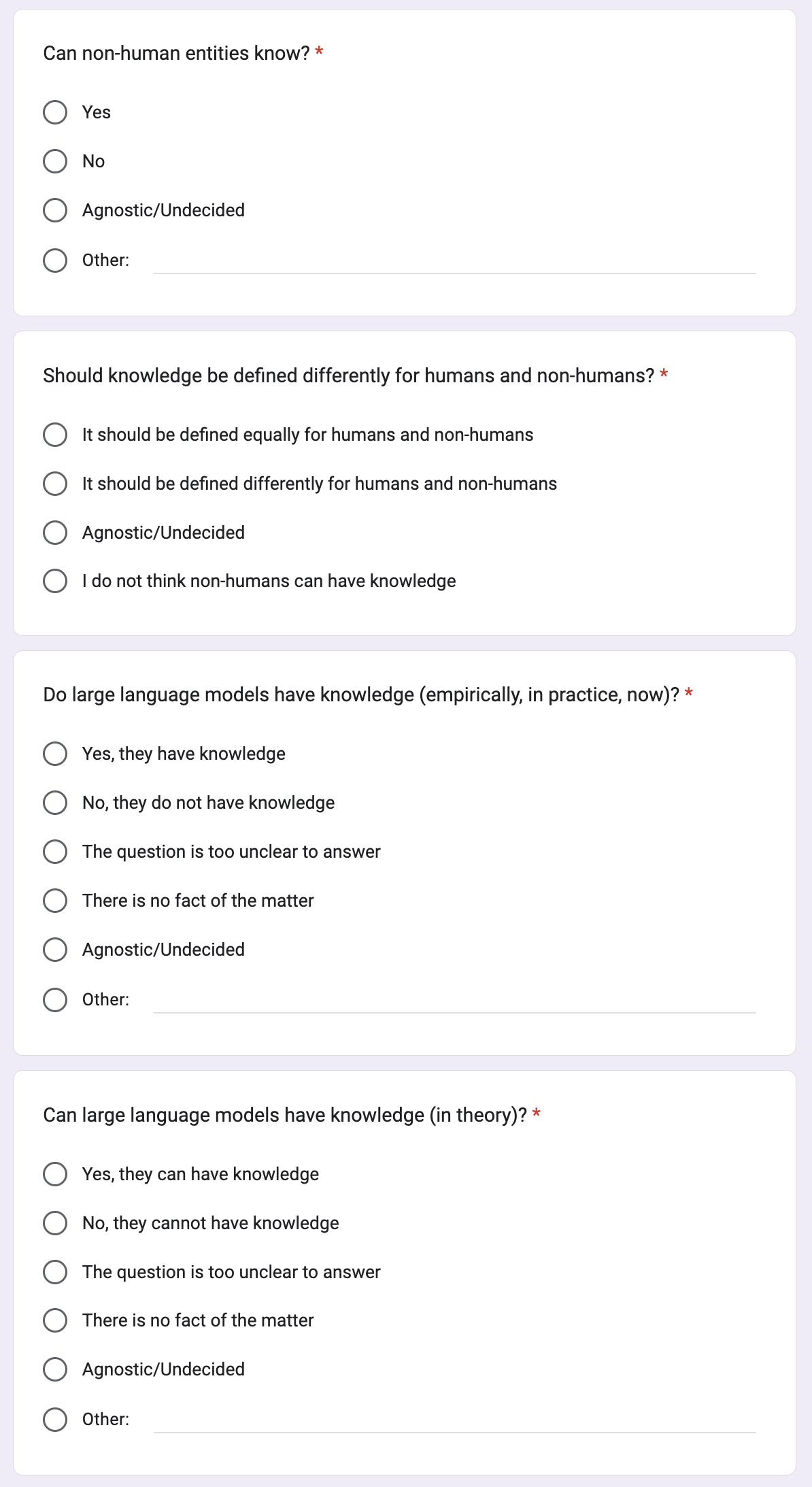}
    \caption{Final part of questions from our survey.}
    \label{fig:survey_can_know_qs}
\end{figure}

We first ask general questions about the respondent and their knowledge of language models and epistemology (Figure \ref{fig:survey_intro_qs}). Then, we present an example to motivate the informal knowledge definitions (Figure \ref{fig:survey_scenario}), and we ask to rank the definitions in a Likert scale (Figure \ref{fig:survey_definitions}). Finally, we ask questions related to whether LLMs can be said to know (Figure \ref{fig:survey_can_know_qs}).

\section{Protocols Example}\label{appendix:protocol_example}
See Table \ref{tab:protocol_example}.

{\begin{table*}[t]
\small 
\centering
\begin{tabular}{p{0.3\linewidth} p{0.65\linewidth}}
\toprule
Prompt & Llama-3's Completion \\ \midrule
\rowcolor[gray]{0.9}
What is a mammal? & A mammal is (...) key characteristics of mammals include:

1. Hair or fur: Mammals are covered in hair or fur, which helps to regulate their body temperature and protect their skin. (...)

5. Live birth: Mammals give birth to live young, rather than laying eggs like birds or reptiles. (...) \\
Do mammals lay eggs? & No, mammals do not lay eggs. Mammals are a group of (...) \\
\rowcolor[gray]{0.9}What are platypus? & The platypus! They're one of the most fascinating and unique creatures in the animal kingdom. The platypus (Ornithorhynchus anatinus) is a semi-aquatic mammal that is native to eastern Australia (...) \\
Do platypuses have hair or fur? & Platypuses do not have hair, but they do have a covering of soft, woolly fur. This fur is made up (...) \\
\rowcolor[gray]{0.9}Do platypuses lay eggs? & Yes, platypuses do lay eggs! In fact, they (...) \\
Are platypuses mammals? & Yes, platypuses are indeed mammals! They belong to the order Monotremata, which (...) \\
\rowcolor[gray]{0.9}Are platypus mammals? Please explain step-by-step. &
\begin{lstlisting}
I'd be happy to help clarify whether platypus are mammals or not.

Here's a step-by-step explanation:

1. **Definition of a mammal**: A mammal is a warm-blooded animal that has hair or fur, produces milk to feed its young, and has mammary glands. Mammals also have a specific type of skeleton, brain structure, and dental formula.

2. **Platypus characteristics**: Platypuses have some characteristics that are typical of mammals, such as:
	* They are warm-blooded, meaning they regulate their own body temperature.
	* They have hair or fur, although it's not very visible.
	* They produce milk to feed their young, just like other mammals.

3. **Unique platypus features**: However, platypuses also have some features that are not typical of mammals:
	* They lay eggs instead of giving birth to live young like most other mammals.
	* They have a duck-billed snout and webbed feet, which are more similar to those of reptiles and birds.
	* They have a venomous spur on their hind leg, which is a characteristic of some reptiles.

4. **Classification**: Based on their characteristics, platypuses are classified as mammals, but they are also considered to be part of a unique group called monotremes. Monotremes are mammals that lay eggs instead of giving birth to live young.

5. **Conclusion**: In conclusion, platypuses are indeed mammals, but they are a special type of mammal that has some characteristics that are different from those of other mammals. Their ability to lay eggs and have a venomous spur are just a few examples of their unique features.
\end{lstlisting} \\
\bottomrule
\end{tabular}
\caption{Greedy decoding of Llama-3-8B-Instruct given a prompt. The `(...)' means that more text was generated but omitted here due to space limitations.}
\label{tab:protocol_example}
\end{table*}}

\end{document}